\documentclass[10pt,twocolumn,letterpaper]{article}

\usepackage{cvpr}
\usepackage{times}
\usepackage{epsfig}
\usepackage{graphicx}
\usepackage{amsmath}
\usepackage{amssymb}

\usepackage{epstopdf}
\usepackage{booktabs} 
\usepackage{epsfig}
\usepackage{graphicx}
\usepackage{amsmath}
\usepackage{amssymb}
\usepackage[ruled,vlined]{algorithm2e}
\usepackage{multirow}

\usepackage{helvet,url}
\usepackage{courier}
\usepackage{epsfig}
\usepackage{picinpar}
\usepackage{graphicx}
\usepackage{epstopdf}
\usepackage{amsmath}
\usepackage{amsthm}
\usepackage{amssymb,bm}
\usepackage[ruled,vlined]{algorithm2e}
\usepackage[titletoc]{appendix}
\usepackage{float,caption}
\usepackage{epsfig}
\usepackage{picinpar}
\usepackage{amsmath}
\usepackage{amsthm}
\usepackage{amssymb,bm}
\usepackage[ruled,vlined]{algorithm2e}
\usepackage[titletoc]{appendix}
\usepackage{float,caption,multirow}
\usepackage{subcaption,xcolor}

\newcommand{\defeq}{\mathrel{\mathop:}=}

\newcommand{\bs}{\mathbf}
\newcommand{\mc}{\mathcal}
\newcommand{\mb}{\mathbb}

\usepackage[pagebackref=true,breaklinks=true,letterpaper=true,colorlinks,bookmarks=false]{hyperref}

 \cvprfinalcopy 

\def\cvprPaperID{2628} 

\begin{document}

\title{PFLD: A Practical Facial Landmark Detector}

\author{Xiaojie Guo$^{1}$, Siyuan Li$^{1}$, Jinke Yu$^{1}$, Jiawan Zhang$^{1}$, Jiayi Ma$^{2}$, Lin Ma$^{3}$, Wei Liu$^{3}$, and Haibin Ling$^{4}$\\
$^{1}$Tianjin University $^{2}$Wuhan University  $^{3}$Tencent AI Lab  $^{4}$Temple University}


\maketitle

\begin{abstract}
   Being accurate, efficient, and compact is essential to a facial landmark detector for practical use. To simultaneously consider the three concerns, this paper investigates a neat model with promising detection accuracy under wild environments (\textit{e.g.}, unconstrained pose, expression, lighting, and occlusion conditions) and super real-time speed on a mobile device. More concretely, we customize an end-to-end single stage network associated with acceleration techniques. During the training phase, for each sample, rotation information is estimated for geometrically regularizing landmark localization, which is then NOT involved in the testing phase. A novel loss is designed to, besides considering the geometrical regularization, mitigate the issue of data imbalance by adjusting weights of samples to different states, such as large pose, extreme lighting, and occlusion, in the training set. Extensive experiments are conducted to demonstrate the efficacy of our design and reveal its superior performance over state-of-the-art alternatives on widely-adopted challenging benchmarks, \textit{i.e.}, 300W (including iBUG, LFPW, AFW, HELEN, and XM2VTS) and AFLW. Our model can be merely 2.1Mb of size and reach over 140 fps per face on a mobile phone (Qualcomm ARM 845 processor) with high precision, making it attractive for large-scale or real-time applications. We have made our practical system based on PFLD 0.25X model publicly available at \url{http://sites.google.com/view/xjguo/fld} for encouraging comparisons and improvements from the community.
\end{abstract}

\section{Introduction}

Facial landmark detection \textit{a.k.a.} face alignment aims to automatically localize a group of pre-defined fiducial points (\textit{e.g.}, eye corners, mouth corners, \textit{etc.}) on human faces. As a fundamental component in a variety of face applications, such as face recognition \cite{FR1,FR2} and verification \cite{FV}, as well as face morphing \cite{FF} and editing \cite{FE}, this problem has been drawing much attention from the vision community with a great progress made over the past years.  
However, \emph{developing a practical facial landmark detector remains challenging, as the detection accuracy, processing speed, and model size should all be concerned.}

\begin{figure}[t]
	\begin{center}
		\begin{subfigure}{0.24\linewidth}
			\includegraphics[width=1\linewidth]{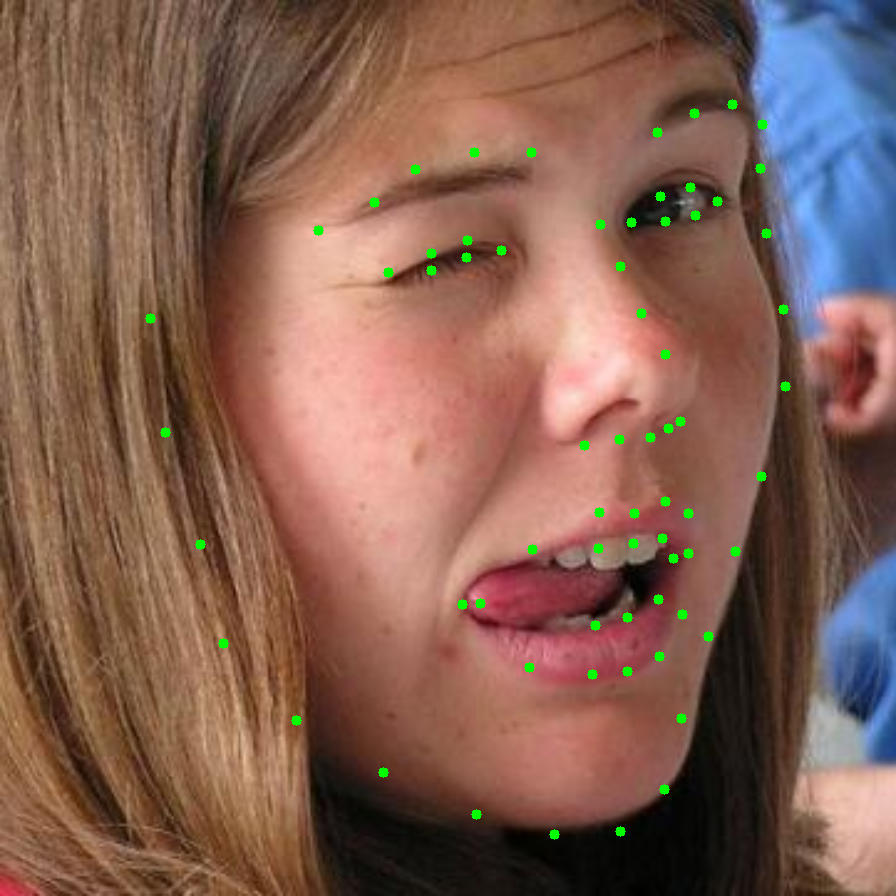}
			\setcounter{subfigure}{0}
			\vspace{-11pt}
		\end{subfigure}
		\begin{subfigure}{0.24\linewidth}
			\includegraphics[width=1\linewidth]{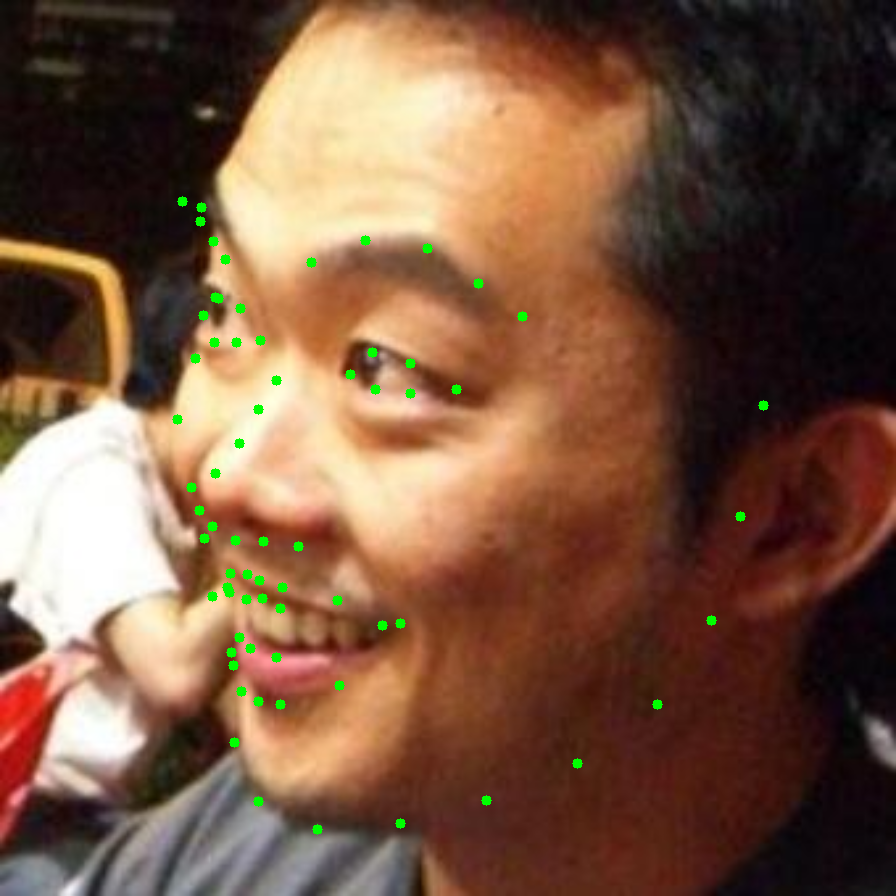}
			\setcounter{subfigure}{0}
			\vspace{-11pt}
		\end{subfigure}
		\begin{subfigure}{0.24\linewidth}
			\includegraphics[width=1\linewidth]{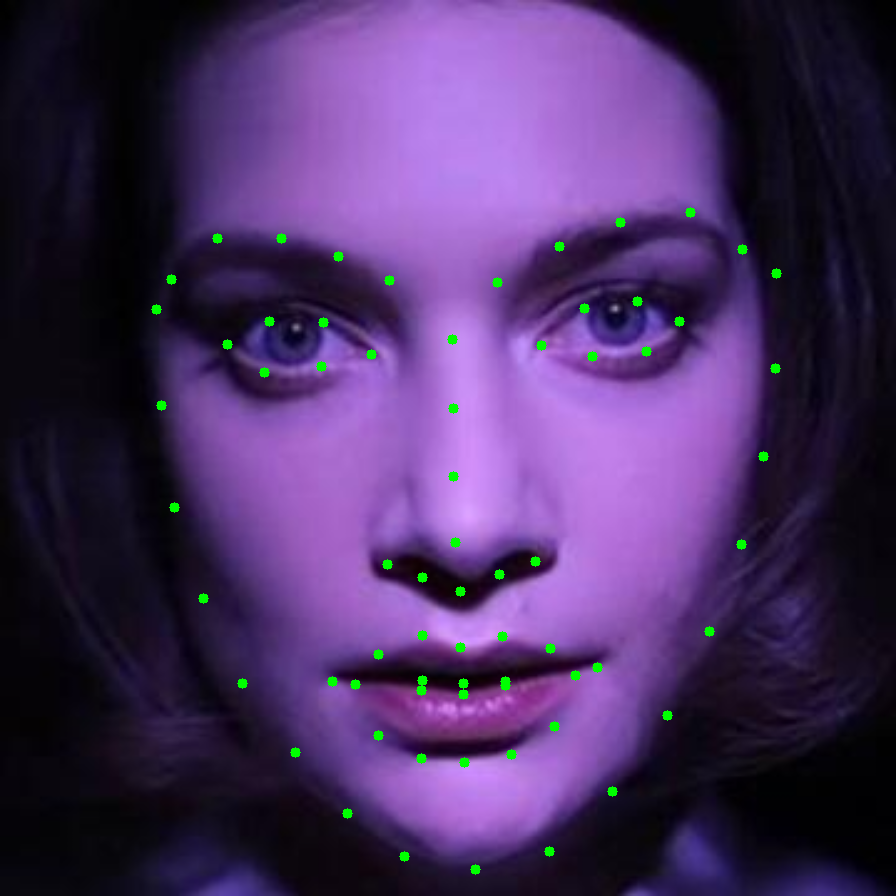}
			\setcounter{subfigure}{0}
			\vspace{-11pt}
		\end{subfigure}
		\begin{subfigure}{0.24\linewidth}
			\includegraphics[width=1\linewidth]{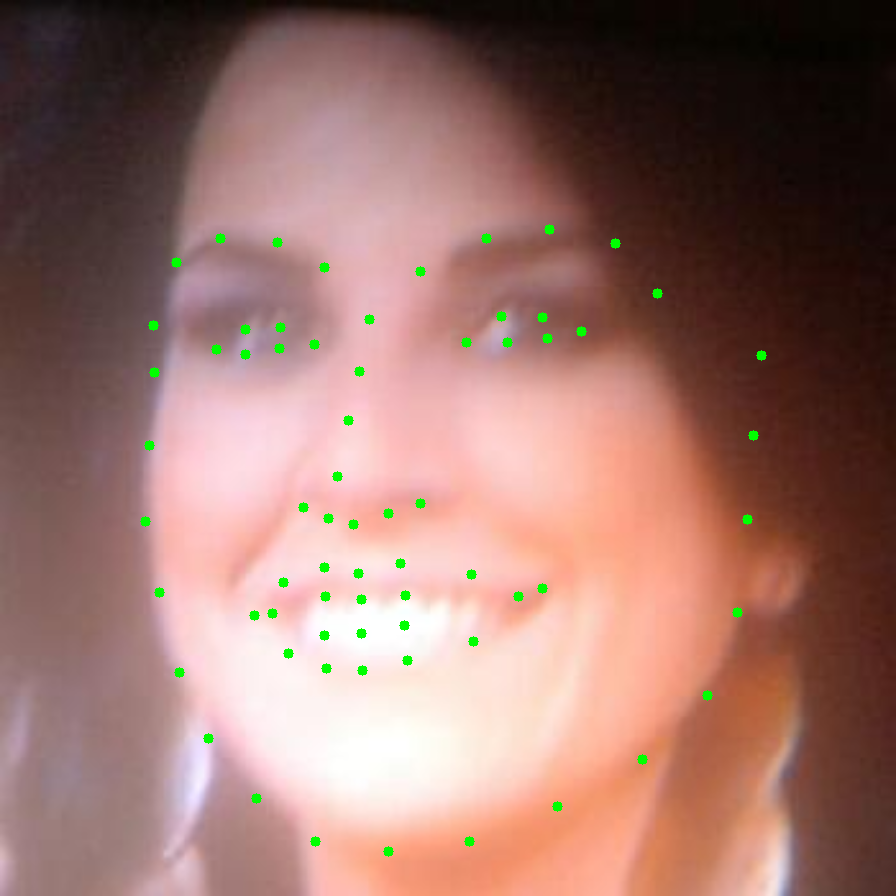}
			\setcounter{subfigure}{0}
			\vspace{-11pt}
		\end{subfigure}
	\begin{subfigure}{0.24\linewidth}
		\includegraphics[width=1\linewidth]{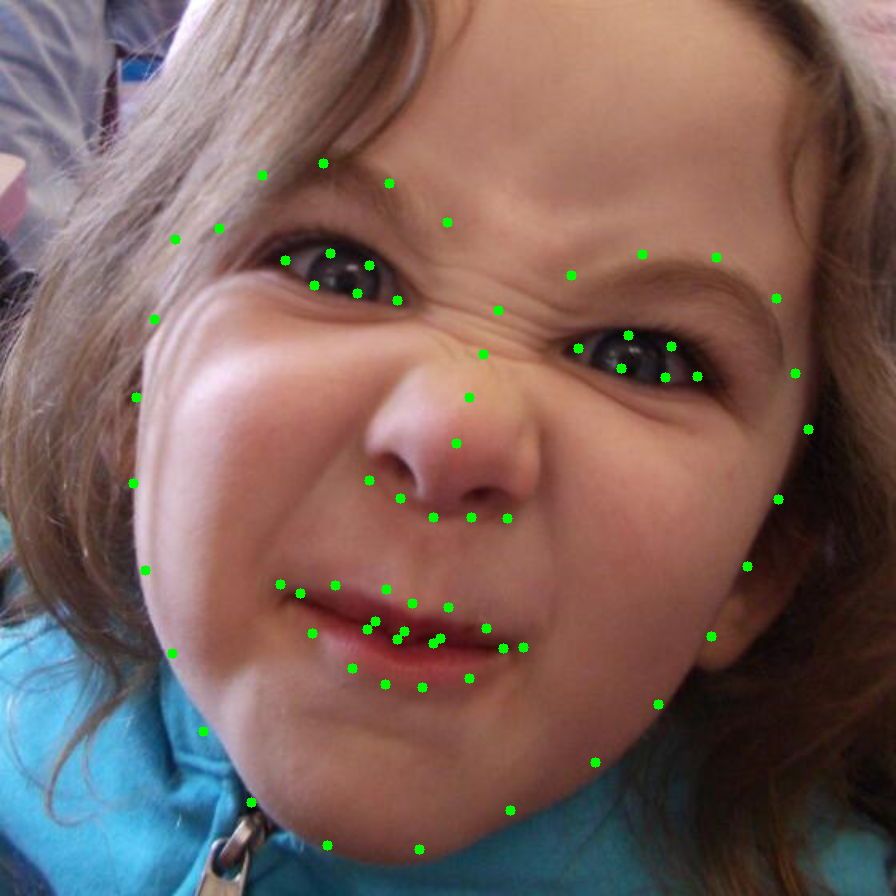}
		\setcounter{subfigure}{0}
		\vspace{-10pt}
	\end{subfigure}
	\begin{subfigure}{0.24\linewidth}
		\includegraphics[width=1\linewidth]{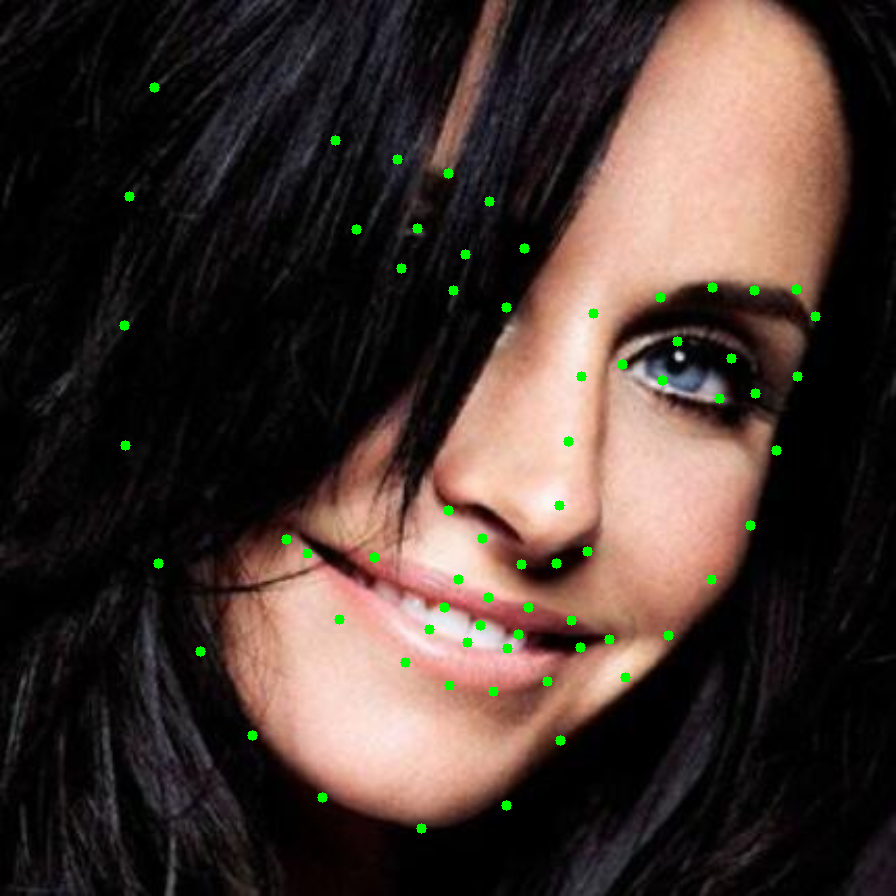}
		\setcounter{subfigure}{0}
		\vspace{-10pt}
	\end{subfigure}
	\begin{subfigure}{0.24\linewidth}
	\includegraphics[width=1\linewidth]{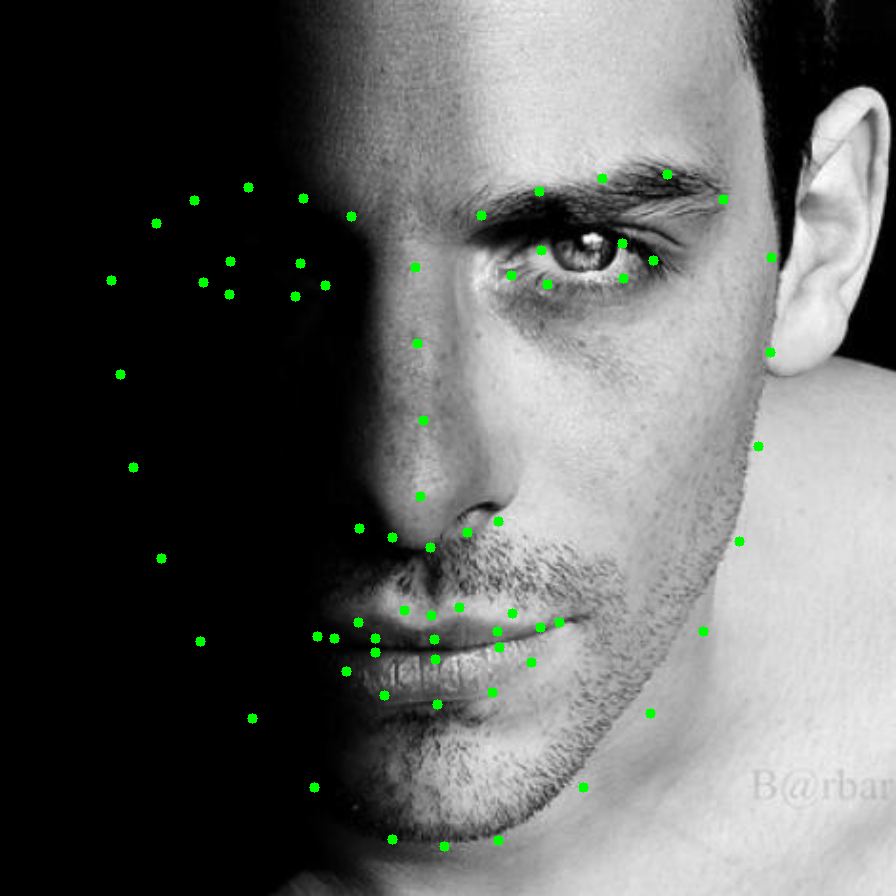}
	\setcounter{subfigure}{0}
	\vspace{-10pt}
\end{subfigure}
	\begin{subfigure}{0.24\linewidth}
		\includegraphics[width=1\linewidth]{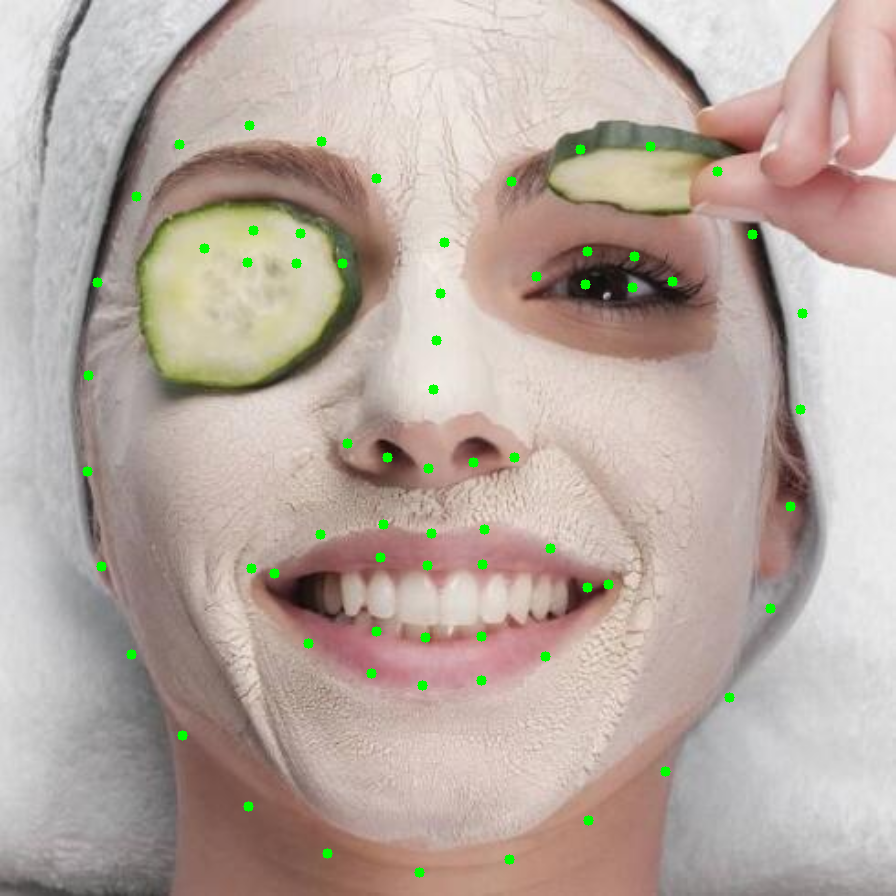}
		\setcounter{subfigure}{0}
		\vspace{-10pt}
	\end{subfigure}
	
	\end{center}
	\vspace{-0pt}
	\caption{Example faces with different poses, expressions, lightings, occlusions, and image qualities. The green markers are detected landmarks via our method. The processing speed achieves over 140 fps on an Android phone with Qualcomm ARM 845 processor.}
	\label{fig:open}
	\vspace{-0pt}
\end{figure}

\emph{Acquiring perfect faces is barely the case in real-world situations.} In other words, human faces are often exposed in under-controlled or even unconstrained environments. The appearance has large variations of poses, expressions and shapes under various lighting conditions, sometimes with partial occlusions. Figure \ref{fig:open} provides several such examples. Besides, sufficient training data for data-driven approaches is also key to model performance. It may be viable to capture several persons' faces under different conditions with balanced consideration though, this collecting manner becomes impractical especially when large-scale data is required to train (deep) models.  Under the circumstances, one often comes across an imbalanced data distribution. The following summarizes issues regarding the landmark detection accuracy into three challenges.\vspace{-0pt}\\
\\ 
\noindent \textbf{Challenge \#1 - Local Variation.} Expression, local extreme lighting (\textit{e.g.}, highlight and shading), and occlusion bring partial changes/interferences onto face images. Landmarks of  some regions may deviate from their normal positions or even disappear.\vspace{-0pt}\\
\\
\textbf{Challenge \#2 - Global Variation.} Pose and imaging quality are two main factors globally affecting the appearance of faces in images, which would result in poor localization of  a (large) fraction of landmarks when the global structure of faces is mis-estimated.\vspace{-0pt}\\
\\
\textbf{Challenge \#3 - Data Imbalance.} It is not uncommon that, in both shallow learning and deep learning, an available dataset exhibits an unequal distribution between its classes/attributes. The imbalance highly likely makes an algorithm/model fail to properly represent the characteristics of the data, thus offering unsatisfactory accuracies across different attributes.\vspace{-7pt}\\
\\ 
{The above challenges considerably increase the difficulty of accurate detection, demanding the detector to be robust.}\\

\emph{With the emergence of portable devices, more and more people prefer to deal with their business or get entertained anytime and anywhere.} Therefore, the challenge below, aside from pursuing high accuracy of detection, should be taken into account.\vspace{-0pt}\\
\\
\noindent \textbf{Challenge \#4 - Model Efficiency. } Another two constraints on applicability are model size and computing requirement. Tasks like robotics, augmented reality, and video chat are expected to be executed in a timely fashion on a platform equipped with limited computation and memory resources \textit{e.g.}, smart phones or embedded products. \vspace{-7pt}\\
\\
This point particularly requires the detector to be of small model size and fast processing speed. Undoubtedly, it is desired to build accurate, efficient, and compact systems for practical landmark detection. 

\subsection{Previous Arts}

Over last decades, a number of classic methods have been proposed in the literature for facial landmark detection. Parameterized appearance models, with \textit{active appearance models} (AAMs) \cite{AAM} and \textit{constrained local models} (CLMs) \cite{CLM} as representatives, accomplish the job through maximizing the confidence of part locations in an image. Specifically, AAMs and its follow-ups \cite{AAMR,AIA,AAMC} attempt to jointly model holistic appearance and shape, while CLMs and variants \cite{CLME,CLMG} instead learn a group of local experts via imposing various shape constraints. In addition, the \textit{tree structure part model} (TSPM) \cite{TSPM} utilizes a deformable part-based model for simultaneous detection, pose estimation, and landmark localization. The methods including \textit{explicit shape regression} (ESR) \cite{ESR} and \textit{supervised descent method} (SDM) \cite{SDM} try to address the problem in a regression manner. The main limitations of these methods are the inferior robustness against difficult cases, expensive computation, and/or high model complexity. A more elaborated review for the classic approaches can be found in \cite{FDLS}. 


Recently, deep learning based strategies have dominated state-of-the-art performances on this task. In what follows, we briefly introduce representative works in this category. Zhang \textit{et al.} \cite{TCDCN} built up a multi-task learning network, called TCDCN, for jointly learning landmark locations and pose attributes. TCDCN, due to its multi-task nature, is difficult to train in practice. An end-to-end recurrent convolutional model for face alignment from coarse to fine was proposed by Trigeorgis \textit{et al.}, termed as MDM \cite{MDM}. Lv \textit{et al.} \cite{TSR} proposed a deep regression architecture with the two-stage re-initialization scheme, namely TSR, which divides a whole face into several parts to boost the detection accuracy. Using pose angles including pitch, yaw, and roll as attributes, \cite{HPE} constructs a network to directly estimate these three angles for helping landmark detection. But the cascaded nature of \cite{HPE} makes it suboptimal in the following landmark detection. \textit{Pose-invariant face alignment} (PIFA for short) proposed by Jourabloo \textit{et al.} \cite{PIFA} estimates the projection matrix from 3D to 2D via deep cascade regressors, which is followed by the work PIFA-CNN \cite{PIFACNN} using a single \textit{convolutional neural network} (CNN). The work in \cite{3DDFA} first models the face depth in a Z-buffer and then fits a 3D model for 2D images. 

Most recently, Kumar and Chellapa designed a single dendritic CNN, named as \textit{pose conditioned dendritic convolution neural network} (PCD-CNN) \cite{PCD-CNN}, which combines a classification network with a second and modular classification network, for improving the detection accuracy. Honari \textit{et al.} designed a network, called \textit{sequential multi-tasking} (SeqMT) net, with an \textit{equivariant landmark transformation} (ELT) loss term \cite{ELT}. In \cite{DCFE}, the authors presented a facial landmark regression method based on a coarse-to-fine \textit{ensemble of regression trees} (ERT) \cite{ERT}. To make the facial landmark detector robust against the intrinsic variance of image styles, Dong \textit{et al.} developed a \textit{style-aggregated network} (SAN) \cite{SAN}, which accompanies the original face images with style-aggregated ones to train the landmark detector. By considering boundary information as the geometric structure of human faces, Wu \textit{et al.} presented a boundary-aware face alignment algorithm, \textit{i.e.} LAB, to improve the detection accuracy. LAB derives face landmarks from boundary lines. By doing so, the ambiguities in the landmark definition can be largely avoided. Other face alignment techniques include \cite{CPM,CFAN,CCL,WingLoss,RAR,RDR}. Though the existing deep learning strategies have made great strides for the task, huge space still exists for improvement especially jointly taking into account the accuracy, efficiency, and model compactness of detectors for practical use.

\subsection{Our Contributions}

The main intention of this work is to show that a good design can save a lot resources with the state-of-the-art performance on the target task. This work develops a \textit{practical facial landmark detector}, denoted as PFLD, with high accuracy against complex situations including unconstrained poses, expressions, lightings, and occlusions.  Compared with the \emph{local variation}, the \emph{global one} deserves more efforts, as it can greatly influence the whole set of landmarks.  To boost the robustness, we employ a branch of network to estimate the \emph{geometric information} for each face sample, and subsequently regularize the landmark localization. Besides, in deep learning, the \emph{data imbalance} issue often limits the performance in accurate detection. For instance, a training set may contain plenty of frontal faces while lacking those with large poses. This would degrade the accuracy when dealing with large pose cases. To address this issue, we advocate to penalize more on errors corresponding to rare training samples than on those to rich ones. Considering the above two concerns, say the \emph{geometric constraint} and the \emph{data imbalance}, a novel loss is designed. 
To enlarge the receptive field and better catch the global structure on faces, a \textit{multi-scale fully-connected} (MS-FC) layer is added for precisely localizing landmarks in images. As for the \emph{processing speed} and \emph{model compactness}, we build the backbone network of our PFLD using MobileNet blocks \cite{MobileV1,MobileV2}.  In experiments, we evaluate the efficacy of our design, and demonstrate its superior performance over other state-of-the-art alternatives on two widely-adopted challenging datasets including 300W \cite{300W} and AFLW \cite{AFLW}. Our model can be adjusted to merely 2.1Mb of size and achieve over 140 fps per face on a mobile phone. All the above merits make our PFLD attractive for practical use. We have released our practical system based on PFLD 0.25X model at \url{http://sites.google.com/view/xjguo/fld} for encouraging comparisons and improvements from the community.

\section{Methodology}
Against the aforementioned challenges, effective measures need to be taken. In this section, we first focus on the design of loss function, which simultaneously takes care of Challenges \#1, \#2, and \#3. Then, we detail our architecture. The whole deep network consists of a backbone subnet for predicting landmark coordinates, which specifically considers Challenge \#4, as well as an auxiliary one for estimating geometric information. 

\subsection{Loss Function}
The quality of training greatly depends on the design of loss function, especially when the scale of training data is not sufficiently large. For penalizing errors between ground-truth landmarks $\bs{X}\defeq[\bs{x}_1,...,\bs{x}_N]\in\mb{R}^{2\times N}$ and predicted ones  $\bs{Y}\defeq[\bs{y}_1,...,\bs{y}_N]\in\mb{R}^{2\times N}$, the simplest losses arguably go to $\ell_2$ and $\ell_1$ losses. However, equally measuring the differences of landmark pairs is not so wise, without considering geometric/structural information. For instance, given a pair of $\bs{x}_i$ and $\bs{y}_i$ with their deviation $\bs{d}_i\defeq\bs{x}_i-\bs{y}_i$ in the image space, if two projections (poses with respect to a camera) are applied from 3D real face to 2D image, the intrinsic distances on the real face could be significantly different. Hence, \emph{integrating geometric information into penalization is helpful to mitigating this issue.} For face images, the global geometric status - 3D pose - is sufficient to determine the manner of projection. Formally, let $\bs{X}$ denote the concerned location of 2D landmarks, which is a projection of 3D face landmarks, \textit{i.e.} $\bs{U}\in\mb{R}^{4\times N}$, each column of which corresponds to a 3D location $[u_i,v_i,z_i,1]^T$. By assuming a weak perspective model as \cite{PIFA}, a $2\times 4$ projection matrix $\bs{P}$ can connect $\bs{U}$  and $\bs{X}$ via $\bs{X}=\bs{P}\bs{U}$. This projection matrix has six degrees of freedom including yaw, roll, pitch, scale, and 2D translation. In this work, the faces are supposed to be well detected, centralized, and normalized\footnote{In our practical system, the face detector \cite{MTCNN} is employed.}. And local variation like expression barely affects the projection. This is to say, three degrees of freedom including scale and 2D translation can be reduced, and thus only three Euler angles (yaw, roll, and pitch) are needed to be estimated. 

 Moreover, in deep learning, data imbalance is another issue often limiting the performance in accurate detection. For example, a training set may contain a large number of frontal faces while lacking those with large poses. Without extra tricks, it is almost sure that the model trained by such a training set is unable to handle large pose cases well. Under the circumstances, ``equally" penalizing each sample makes it unequal instead. \emph{To address this issue, we advocate to penalize more on errors corresponding to rare training samples than on those to rich ones.} 

\begin{figure*}[t] 
	\centering
	\includegraphics[width=0.85\linewidth]{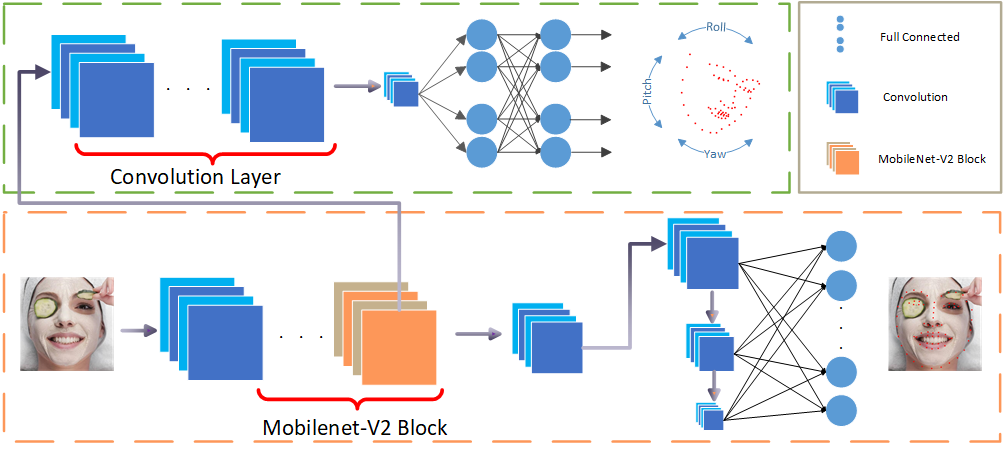}
	\caption{The illustration of our architecture. The whole network consists of two subnets, including the backbone network (lower branch) for predicting landmark coordinates and the auxiliary one (upper branch) for estimating geometric information.}
	\label{fig:network} 
	\vspace{0pt}
\end{figure*}

Mathematically, the loss can be written in the following general form:
\begin{equation}
\mc{L}\defeq \frac{1}{M}\sum_{m=1}^M\sum_{n=1}^N \gamma_n\|\bs{d}_n^m\|,
\label{eq:gl}
\end{equation}
where $\|\cdot\|$ designates a certain metric to measure the distance/error of the $n$-th landmark of the $m$-th input. $N$ is the pre-defined number of landmarks per face to detect. $M$ denotes the number of training images in each process. Given the metric used (\textit{e.g.}, $\ell_2$ in this work), the weight $\gamma_n$ plays a key role. 
Consolidating the aforementioned concerns, say the geometric constraint and the data imbalance, a novel loss is designed as follows:
\begin{equation}
\mc{L}\defeq \frac{1}{M}\sum_{m=1}^M\sum_{n=1}^N\left(\sum_{c=1}^C\omega_n^c\sum_{k=1}^K(1-\cos\theta_n^k)\right)\|\bs{d}_n^m\|_2^2.
\label{eq:ol}
\end{equation}
It is easy to obtain that $\sum_{c=1}^C\omega_n^c\sum_{k=1}^K(1-\cos\theta_n^k)$ in Eq. \eqref{eq:ol} acts as $\gamma_n$ in Eq. \eqref{eq:gl}. Let us here take a close look at the loss. In which,
$\theta^1$, $\theta^2$, and $\theta^3$ (K=3) represent the angles of deviation between the ground-truth and estimated yaw, pitch, and roll angles. Clearly, as the deviation angle increases, the penalization goes up.  In addition, we categorize a sample into one or multiple attribute classes including profile-face, frontal-face, head-up, head-down, expression, and occlusion. The weighting parameter $\omega_n^c$ is adjusted according to the fraction of samples belonging to class $c$ (this work simply adopts the reciprocal of fraction).  For instance, if disabling the geometry and data imbalance functionalities, our loss degenerates to a simple $\ell_2$ loss. No matter whether the 3D pose and/or the data imbalance bother(s) the training or not, our loss can handle the local variation by its distance measurement. 

\emph{Although, in the literature, several works have considered the 3D pose information to improve the performance, our loss has following merits: 1) it plays in a coupled way between 3D pose estimation and 2D distance measurement, which is much more reasonable than simply adding two concerns \cite{PIFA,PIFACNN}; 2) it is intuitive and easy to be computed both forward and backward, comparing with \cite{PCD-CNN}; and 3) it makes the network work in a single-stage manner instead of cascaded \cite{HPE,PIFA}, which improves the optimality.} 
We here notice that the variable $\bs{d}_n^m$  comes from the backbone net, while  $\theta_n^k$  from the auxiliary one, which are coupled/connected by the loss  in Eq. \eqref{eq:ol}. In the next two subsections, we detail our network, which is schematically illustrated in Fig. \ref{fig:network}.


\subsection{Backbone Network}


Similar to other CNN based models, we employ several convolutional layers to extract features and predict landmarks. Considering that human faces are of strong global structures, like symmetry and spacial relationships among eyes, mouth, nose, \textit{etc.}, such global structures could help localize landmarks more precisely. Therefore, instead of single scale feature maps, we extend them into multi-scale maps. The extension is finished via executing convolution operations with strides, which enlarges the receptive field. Then we perform the final prediction through fully connecting the multi-scale feature maps. The detailed configuration of the backbone subnet is summarized in Table \ref{tab:backbone}. From the perspective of architecture, the backbone net is simple. Our primary intention is to verify that, associated with our novel loss and the auxiliary subnet (discussed in the next subsection), even a very simple architecture can achieve state-of-the-art performance.

\begin{table}[t]
	\begin{center}
		\resizebox{0.48\textwidth}{!}{
			\begin{tabular}{ c| c | c | c | c | c }
				\hline
				\textbf{Input} & \textbf{Operator } & \textbf{$t$} & \textbf{$c$} & \textbf{$n$} & \textbf{$s$}\\
				\hline
				\hline
				$112^2\times 3$ & Conv$3\times 3$ & - & 64 & 1 & 2\\
				$56^2\times 64$ & Depthwise Conv$3\times 3$ & - & 64 & 1 & 1\\
				$56^2\times 64$ & Bottleneck & 2 & 64 & 5 & 2\\
				$28^2\times 64$ & Bottleneck & 2 & 128 & 1 & 2\\
				$14^2\times 128$ & Bottleneck & 4 & 128 & 6 & 1\\
				$14^2\times 128$ & Bottleneck & 2 & 16 & 1 & 1\\
				\hline
				(S1) {$14^2\times 16$ }& {Conv$3\times 3$} & - & 32 & 1 & 2\\
				(S2) $7^2\times 32$ & Conv$7\times 7$ & - & 128 & 1 & 1\\
				(S3) $1^2\times 128$ & - & - & 128 & 1 & -\\
				\hline
				S1, S2, S3& Full Connection & - & 136 & 1 & -\\
				\hline
			\end{tabular}
		}
		\vspace{-0pt}
	\end{center}
	\caption{The backbone net configuration. Each line represents a sequence of identical layers, repeating	$n$ times. All layers in the same sequence have the same number $c$ of output channels. The first layer of each sequence has a stride $s$. The expansion factor $t$ is always applied to the input size.}
	\vspace{-0pt}
	\label{tab:backbone}
\end{table}

\begin{table}[t]
	\begin{center}
		\begin{tabular}{ c| c | c | c  }
			\hline
			\textbf{Input} & \textbf{Operator } & \textbf{$c$} &  \textbf{$s$}\\
			\hline
			\hline
			$28^2\times 64$ &  Conv$3\times 3$ & 128 & 2\\
			$14^2\times 128$ & Conv$3\times 3$ & 128 & 1\\
			$14^2\times 128$ & Conv$3\times 3$ & 32 & 2\\
			$7^2\times 32$ & Conv$7\times 7$ & 128 & 1\\
			\hline
			$1^2\times 128$& Full Connection & 32 & 1\\
			$1^2\times 32$& Full Connection & 3 & -\\
			\hline
		\end{tabular}
	\end{center}
	\vspace{-0pt}
	\caption{The auxiliary net configuration. As the auxiliary branch is no longer needed in the testing, we do not apply the MobileNet techniques in our implementation.}
	\vspace{-0pt}
	\label{tab:aux}
\end{table}

The backbone network is the bottleneck in terms of processing speed and model size, as in the testing only this branch is involved. Thus, it is critical to make it fast and compact. Over the last years, several strategies including ShuffleNet \cite{Shufflenet}, Binarization \cite{Binary}, and MobileNet \cite{MobileV1} have been investigated to speed up networks. 
Due to the satisfactory performance of MobileNet techniques (depthwise separable convolutions, linear bottlenecks, and inverted residuals) \cite{MobileV1,MobileV2}, we replace the traditional convolution operations with the MobileNet blocks. By doing so, the computational load of our backbone network is significantly reduced and the speed is thus accelerated. 
In addition, our network can be compressed by adjusting the width parameter of MobileNets according to demand from users, for making the model smaller and faster. This operation is based on the observation and assumption that a large amount of individual feature channels of a deep convolutional layer could lie in a lower-dimensional manifold. Thus, it is highly possible to reduce the number of feature maps without (obvious) accuracy degradation. We will show in experiments, losing 80\% of the model size can still provide promising accuracy of detection. This again corroborates that a well-designed simple/small architecture can perform sufficiently well on the task of facial landmark detection. It is worth to mention that the quantization techniques are totally compatible with ShuffleNet and MobileNet, which means the size of our model can be further reduced by quantization.

\subsection{Auxiliary Network}

It has been verified by previous works \cite{3DDFA,PIFA,PCD-CNN,LAB} that a proper auxiliary constraint is beneficial to making the landmark localization stable and robust. Our auxiliary network plays this role. Different from the previous methods, like \cite{PIFA} learning the 3D to 2D projection matrix, \cite{PCD-CNN} discovering the dendritic structure of parts, and \cite{LAB} employing boundary lines, our intention is to estimate the 3D rotation information including yaw, pitch, and roll angles. Having these three Euler angles, the pose of head can be determined. 

One may wonder that \emph{given predicted and ground-truth landmarks, why not directly compute the Euler angles from them?} Technically, it is feasible. However, the landmark prediction may be too inaccurate especially at the beginning of training, which consequently results in a low-quality estimation of the angles. This could drag the training into dilemmas, like over-penalization and slow convergence. To decouple the estimation of rotation information from  landmark localization, we bring the auxiliary subnet.  

It is worth mentioning that DeTone \textit{et al.} \cite{HomoE} proposed a deep network for estimating the homography between two related images. The yaw, roll, and pitch angles can be calculated from the estimated homography matrix. But for our task, we do not have a frontal face with respect to each training sample. Intriguingly, our auxiliary net can output the target angles without a requirement of frontal faces as input. The reason is that our task is specific to human faces that are of strong regularity and structure from the frontal view. In addition, the factors such as expressions and lightings barely affect the pose. Thus, an identical average frontal face can be considered available for different persons. In other words, there is NO extra annotation used for computing the Euler angles. The following is our way to calculate them: 1) predefine ONE standard face (averaged over a bunch of frontal faces) and fix 11 landmarks on the dominant face plane as references for ALL of training faces; 2) use the corresponding 11 landmarks of each face and the reference ones to estimate the rotation matrix; and 3) compute the Euler angles from the rotation matrix. For accuracy, the angles may not be exact for each face, as the averaged face is used for all the faces. Even though, they are sufficiently accurate for our task as verified later in experiments. Table \ref{tab:aux} provides the configuration of our proposed auxiliary network. Please notice that the input of the auxiliary net is from the $4$-th block of the backbone net (see Table \ref{tab:backbone}).

\subsection{Implementation Details} 

During training, all faces are cropped and resized into $112\times 112$ according to given bounding boxes for pre-processing. We implement the network via the Kera framework, using the  batch size of 256, and employ the Adam technique for optimization with the weight decay of $10^{-6}$ and momentum of $0.9$. The maximum number of iterations is 64K, and the learning rate is fixed to $10^{-4}$ throughout the training. The entire network is trained on a Nvidia GTX 1080Ti GPU. For 300W, we augment the training data by flipping each sample and rotating them from $-30^\circ$ to $30^\circ$ with $5^\circ$ interval. Further, each sample has a region of 20\% face size randomly occluded. While for AFLW, we feed the original training set into the network without any data augmentation. In the testing, only the backbone network is involved, which is efficient. 
\section{Experimental Evaluation}
\subsection{Experimental Settings}
\textbf{Datasets.} To evaluate the performance of our proposed PFLD, we conduct experiments on two widely-adopted challenging datasets, say 300W \cite{300W} and AFLW \cite{AFLW}.

\noindent\textit{300W.} This dataset annotates five face datasets including LFPW, AFW, HELEN, XM2VTS and IBUG, with 68 landmarks. We follow \cite{SAN,LAB,PCD-CNN} to utilize 3,148 images for training and 689 images for testing. The testing images are divided into two subsets, say the common subset formed by 554 images from LFPW and HELEN, and the challenging subset by 135 images from IBUG. The common and the challenging subsets form the full testing set.

{\noindent\textit{AFLW}.} This dataset consists of 24,386 in-the-wild human faces, which are obtained from Flicker with extreme poses, expressions and occlusions. The faces are with head pose ranging from $0^\circ$ to $120^\circ$ for yaw, and upto $90^\circ$ for pitch and roll, respectively. AFLW offers at most 21 landmarks for each face. We use 20,000 images and 4,386 images for training and testing, respectively.\\

{\textbf{Competitors}.} The compared approaches include classic and recently proposed deep learning based schemes, which are RCPR (ICCV'13) \cite{RCPR}, SDM (CVPR'13) \cite{SDM},  CFAN (ECCV'14) \cite{CFAN}, CCNF (ECCV'14) \cite{CCNF}, ESR (IJCV'14) \cite{ESR}, ERT (CVPR'14) \cite{ERT}, LBF (CVPR'14) \cite{LBF}, TCDCN (ECCV'14) \cite{TCDCN}, CFSS (CVPR'15) \cite{CFSS}, 3DDFA (CVPR'16) \cite{3DDFA}, MDM (CVPR'16) \cite{MDM}, RAR (ECCV'16) \cite{RAR}, CPM (CVPR'16) \cite{CPM},  DVLN (CVPRW'17) \cite{DVLN}, TSR (CVPR'17) \cite{TSR}, Binary-CNN (ICCV'17) \cite{Binary}, PIFA-CNN (ICCV'17) \cite{PIFACNN}, RDR (CVPR'17) \cite{RDR}, DCFE (ECCV'18) \cite{DCFE}, SeqMT (CVPR'18) \cite{ELT}, PCD-CNN (CVPR'18) \cite{PCD-CNN}, SAN (CVPR'18) \cite{SAN} and LAB (CVPR'18) \cite{LAB}. \\

{\textbf{Evaluation Metrics}.} Following most previous works \cite{ESR,LAB,SAN,PCD-CNN}, \textit{normalized mean error} (NME) is employed to measure the accuracy, which averages normalized errors over all annotated landmarks. For 300W, we report the results using two normalizing factors. One adopts the eye-center-distance as the \textit{inter-pupil} normalizing factor, while the other is normalized by the outer-eye-corner distance denoted as \textit{inter-ocular}. For ALFW, due to various profile faces, we follow \cite{PCD-CNN,SAN,LAB} to normalize the obtained error by the ground-truth bounding box size over all visible landmarks. The \textit{cumulative error
distribution} (CED) curve is also used to compare the methods. Besides the accuracy, the processing speed and model size are also compared.

%


\begin{table*}[t]
	\begin{center}
		\resizebox{1\textwidth}{!}{
			\begin{tabular}{ c|| c | c |  c | c | c}
				\hline
				\textbf{Model} & {SDM} \cite{SDM}  & {SAN} \cite{SAN}  & {LAB} \cite{LAB} & \textbf{PFLD 0.25X} & \textbf{PFLD 1X}\\
				\hline
				\hline
				\textbf{Size (Mb)} & 10.1   &  270.5+528 & 50.7 & \textbf{2.1} & \textbf{12.5}\\
				\textbf{Speed} & 16ms (C)  &  343ms(G) & 2.6s(C)/60ms(G*) & \textbf{1.2ms(C)/1.2ms(G)/7ms(A)}& \textbf{6.1ms(C)/3.5ms(G)/26.4ms(A)} \\
				\hline
			\end{tabular}
		}
	\end{center}
	\caption{Comparison in terms of model size and processing speed. }
	\label{tab:size}
\end{table*}

\subsection{Experimental Results}
\begin{table}[t]
	\begin{center}
		\begin{tabular}{ c|| c c c }
			\hline
			\textbf{Method} & \textbf{Common} & \textbf{Challenging}  & \textbf{Fullset}\\
			\hline
			\multicolumn{4}{c}{ \text{Inter-pupil Normalization} (IPN)}\\
			\hline 
			{RCPR} \cite{RCPR} & 6.18 & 17.26 &8.35\\
			{CFAN} \cite{CFAN}& 5.50 & 16.78 &7.69\\
			{ESR} \cite{ESR} & 5.28 & 17.00 &7.58\\
			{SDM} \cite{SDM} & 5.57 & 15.40 &7.50\\
			{LBF} \cite{LBF}  & 4.95 & 11.98 &6.32\\
			{CFSS} \cite{CFSS} & 4.73 & 9.98 &5.76\\
			{3DDFA} \cite{3DDFA} & 6.15 & 10.59 &7.01\\
			{TCDCN} \cite{TCDCN} & 4.80 & 8.60 &5.54\\
			{MDM}  \cite{MDM}& 4.83 & 10.14 &5.88\\
			{SeqMT}  \cite{ELT}& 4.84 & 9.93 &5.74\\
			{RAR} \cite{RAR} & 4.12 & 8.35 &4.94\\
			{DVLN} \cite{DVLN} & 3.94 & 7.62 &4.66\\
			{CPM} \cite{CPM} & 3.39 & 8.14 &4.36\\
			{DCFE} \cite{DCFE} & 3.83 & 7.54 &4.55\\
			{TSR} \cite{TSR} & 4.36 & 7.56 &4.99\\
			{LAB} \cite{LAB} & 3.42 & 6.98 &4.12\\
			
			\hline
			\textbf{PFLD 0.25X}  & \textbf{3.38} & \textbf{6.83} &\textbf{4.02}\\
			\textbf{PFLD 1X}  & \textbf{3.32} &\textbf{6.56} &\textbf{3.95}\\
			\textbf{PFLD 1X+}  & \textbf{3.17} & \textbf{6.33} &\textbf{3.76}\\
			\hline
			\multicolumn{4}{c}{ \text{Inter-ocular Normalization} (ION)}\\
			\hline
			{PIFA-CNN} \cite{PIFACNN} & 5.43 & 9.88 &6.30\\
			{RDR} \cite{RDR} & 5.03 & 8.95 &5.80\\
			{PCD-CNN} \cite{PCD-CNN} & 3.67 & 7.62 &4.44\\
			{SAN} \cite{SAN}  & 3.34 & 6.60 &3.98\\
			
			\hline 
			\textbf{PFLD 0.25X}  & \textbf{3.03} & \textbf{5.15} &\textbf{3.45}\\
			\textbf{PFLD 1X}  & \textbf{3.01} & \textbf{5.08} &\textbf{3.40}\\
			\textbf{PFLD 1X+}  & \textbf{2.96} & \textbf{4.98} &\textbf{3.37}\\
			\hline
		\end{tabular}
		\caption{Comparison in normalized mean error on the 300W Common Subset, Challenging Subset, and Fullset.}
			\vspace{-0pt}
	\end{center}
	\label{tab:300w}
\end{table}


\begin{figure}[t]
	\begin{center}
		\includegraphics[width=1\linewidth]{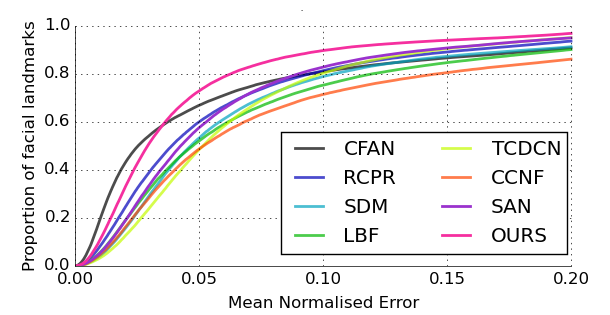}
	\end{center}
\vspace{-0pt}
	\caption{CED curves for the 300W dataset with bounding boxes determined by the 300W face detector.}
	\vspace{-0pt}
	\label{fig:CED}
\end{figure}

\textbf{Detection Accuracy.} We first compare our PFLD against the state-of-the-art methods on 300W dataset. The results are given in Table 4. Three versions of our model including PFLD 0.25X, PFLD 1X and PFLD 1X+ are reported. PFLD 1X and PFLD 0.25X respectively stand for the entire model and the compressed one by setting the width parameter (of MobileNet blocks) to 0.25, both trained using 300W training data only, while PFLD 1X+ represents the entire model additionally pre-trained on the WFLW dataset \cite{WFLW}.  From the numerical results in Table 3, we can observe that our PFLD 1X significantly outperforms previous methods, especially on the challenging subset. Though the performance of PFLD 0.25X is slightly behind that of PFLD 1X, it still achieves better results than the other competitors including most recently proposed LAB \cite{LAB}, SAN \cite{SAN} and PCD-CNN \cite{PCD-CNN}. This comparison is evident to show that PFLD 0.25X is a good trade-off in practice, which cuts about 80\% model size without sacrificing much in accuracy. It also corroborates the assumption that a large number of feature channels of a deep learning convolutional layer could lie in a lower-dimensional manifold. We will see shortly that the speed of PFLD 0.25X is also largely accelerated compared with PFLD 1X. As for PFLD 1X+, it further enlarges the margin of precision to the others.  This indicates that there is space for our network to achieve further improvement by feeding in more training data.

Moreover, we provide CED curves to evaluate the accuracy difference in Fig. \ref{fig:CED}. From a more practical perspective, different from the previous comparison (the ground-truth bounding boxes of faces are given, which are constructed according to ground-truth landmarks), the faces in the testing set are detected by 300W detector for all the involved competitors in this experiment. The performance of some compared methods might be degraded compared with using GT bounding boxes, such as SAN, which reflects the stability of the landmark detector with respect to the face detector. From the curves, we can see that PFLD can outperform the others by a large margin. 

We further evaluate the performance difference among different methods on AFLW. Table 5 reports the NME results obtained by the competitors. As can be observed from the table,  the methods including TSR, CPM, SAN and our PFLDs significantly outperform the rest competing approaches. Among TSRM CPM, SAN and PFLDs, our PFLD 1X achieves the best accuracy (NME 1.88) followed by SAN (NME1.91). The third place is taken by our PFLD 0.25X with competitive NME 2.07. We again emphasize that the model size and processing speed of PFLD 0.25X are greatly superior over those of SAN, please see Table 3.\\\\

\begin{figure*}[!ht]
	\begin{center}
			\includegraphics[width=1\linewidth]{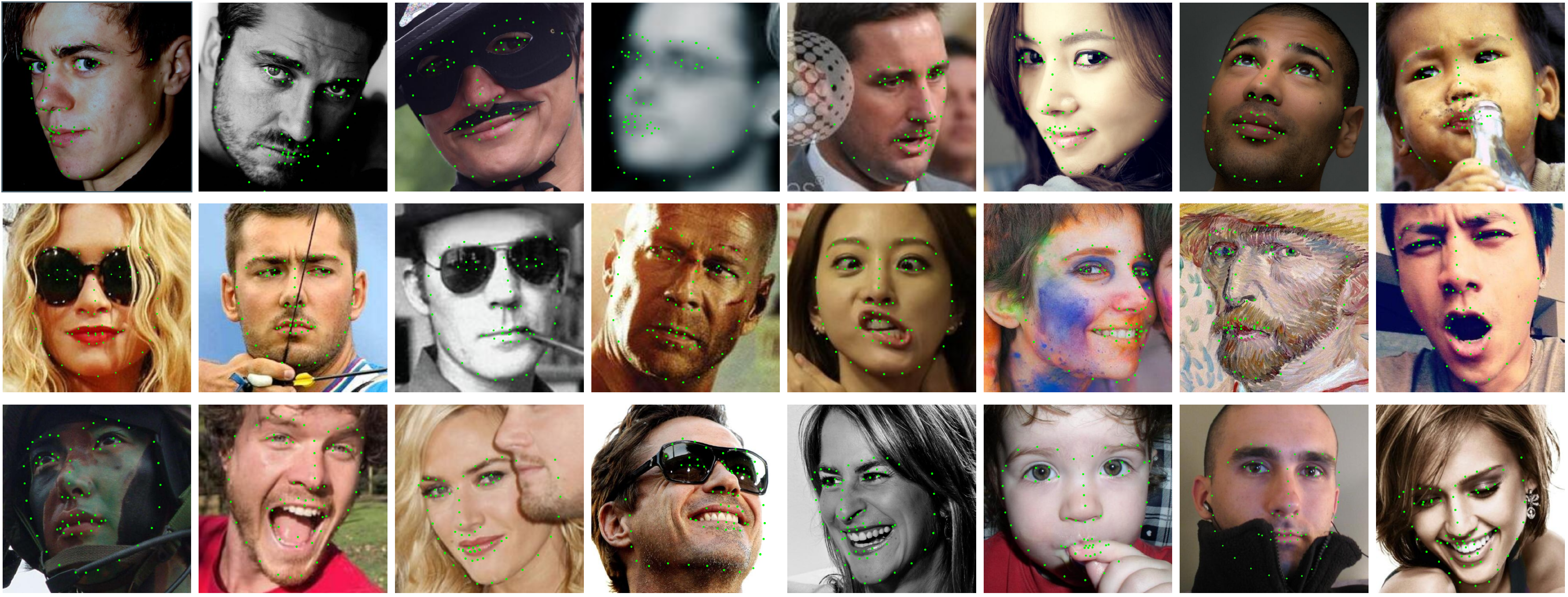}
	\end{center}
	\caption{Qualitative results on several challenging faces by our PFLD 0.25X. We can observe that even with extreme lighting, expression, occlusion, and blur interferences, PFLD 0.25X can obtain visually pleasant results.}
	\label{fig:visual}

\end{figure*}

\textbf{Model Size.} Table 3 compares our PFLDs with some classic and recently proposed deep learning methods in terms of model size and processing speed. As for model size, our PFLD 0.25X is merely 2.1Mb,  saving more than 10Mb from PFLD 1X. PFLD 0.25X is much smaller than the other models including SDM 10.1Mb, LAB 50.7Mb and SAN about 800Mb (containing two VGG-based subnets 270.5Mb+528Mb). \\

\begin{table*}[t]
	\begin{center}
		\begin{tabular}{ c|| c | c | c | c | c | c | c }
			\hline
			\textbf{Method} & {RCPR} \cite{RCPR} & {CDM} \cite{CDM}& {SDM} \cite{SDM} & {ERT} \cite{ERT} & {LBF} \cite{LBF} & {CFSS} \cite{CFSS} & {CCL} \cite{CCL}\\ 
			\hline
			\hline
			\textbf{AFLW} & 5.43 & 3.73 & 4.05 & 4.35 & 4.25 & 3.92 & 2.72 \\
			\hline
			\hline
			\textbf{Method} & {Binary-CNN \cite{Binary} } & {PCD-CNN \cite{PCD-CNN}} & {TSR} \cite{TSR} & {CPM} \cite{CPM}& {SAN} \cite{SAN}& \textbf{PFLD 0.25X} & \textbf{PFLD 1X} \\
			\hline
			\hline
			\textbf{AFLW} & 2.85 & 2.40  & 2.17 & 2.33 & 1.91 & \textbf{2.07} & \textbf{1.88} \\
			\hline
		\end{tabular}
	\end{center}
	\caption{Comparison in normalized mean error on the AFLW-full dataset.}
	\label{tab:aflw}
\end{table*}

\textbf{Processing Speed.} Further, we evaluate the efficiency of each algorithm on an i7-6700K CPU (denoted as C) and a Nvidia GTX 1080Ti GPU (denoted as G) unless otherwise stated. Since only the CPU version of SDM \cite{SDM} and the GPU version of SAN \cite{SAN} are publicly available, we only report the elapsed CPU time and GPU time for them respectively. As for LAB \cite{LAB}, only the CPU version can be downloaded from its project page. Nevertheless, in the paper \cite{LAB}, the authors stated that their algorithm costs about 60ms on a TITAN X GPU (denoted as G*).  As can be seen from the comparison, our PFLD 0.25X and PFLD 1X are remarkably faster than the others in both CPU and GPU times. Please note that the CPU time of LAB is in seconds instead of in milliseconds. The proposed PFLD 0.25X spends the same time (1.2ms) on CPU and GPU, this is because most of time comes from I/O operations. Moreover, PFLD 1X takes about 5 times in CPU and 3 times in GPU of PFLD 0.25X. Even though, PFLD 1X still performs much faster that the others. In addition, for PFLD 0.25X and PFLD 1X, we also perform a test on a Qualcomm ARM 845 processor (denoted as A in the table). Our PFLD 0.25X spends 7ms per face (over 140 fps) while PFLD 1X costs 26.4ms per face (over 37 fps). \\

\textbf{Ablation Study.} To validate the advantages of our loss, we further carry out ablation study on both of 300W and AFLW. Two typical losses including $\ell_2$ and $\ell_1$ are involved. As shown in Table 6, the difference between $\ell_2$ and $\ell_1$ losses is not very obvious, which obtain [4.40 vs. 4.35] in terms of IPN on 300W and [2.33 vs. 2.31] in NME on AFLW, respectively. We note that our base loss is $\ell_2$. Three settings are considered: $\ell_2$ with the geometric constraint only ($\omega^c=1$, denoted as ours w/o $\omega$), $\ell_2$ with the weighting strategy only ($\theta^k=0$, disabling the auxiliary network, denoted as ours w/o $\theta$), and $\ell_2$ with both the geometric constraint and the weighting strategy (denoted as ours). From the numerical results, we see that both ours w/o $\theta$ and ours w/o $\omega$ respectively improve the base $\ell_2$, by relative 4.1\% (IPN 4.22) and 5.9\% (IPN 4.14) on 300W, and relative 4.3\% (NME 2.23) and 7.3\% (NME 2.16) on AFLW. By taking into account both the geometric information and the weighting trick, ours catches relative 10.2\% (IPN 3.95) improvement on 300W and 19.3\% (NME 1.88) on AFLW, respectively. This study verifies the effectiveness of the design of our loss.\\

\begin{figure*}[!t]
	\begin{center}
		\begin{subfigure}{0.85\linewidth}
			\includegraphics[width=\linewidth]{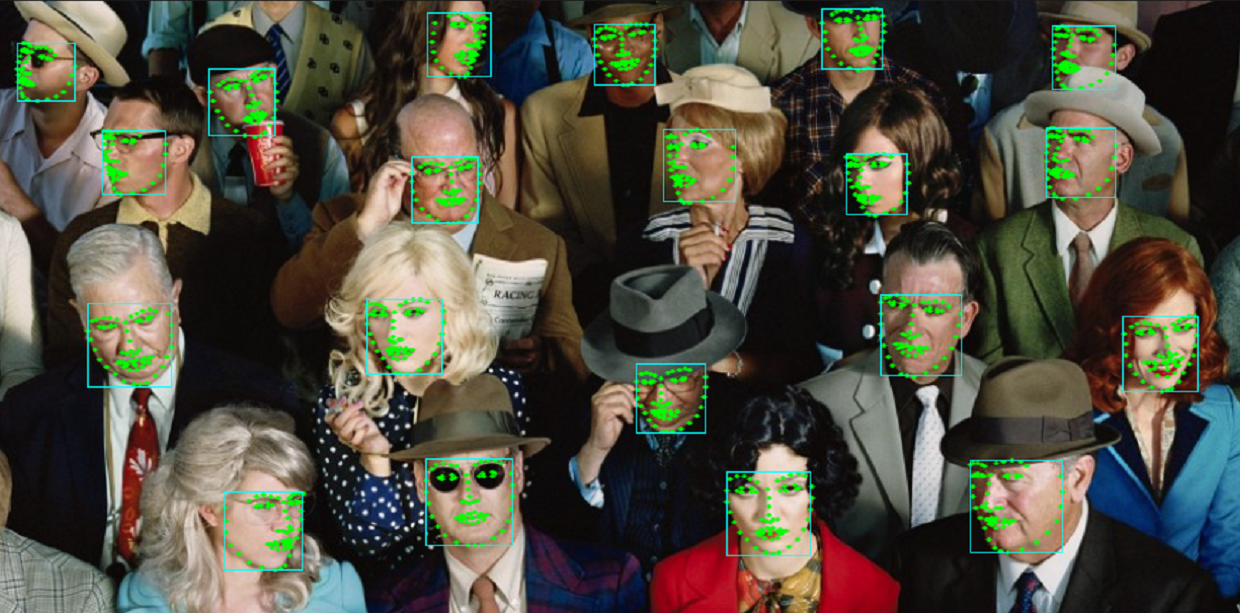}
			\setcounter{subfigure}{0}
			\vspace{-5pt}
		\end{subfigure}
		
		\begin{subfigure}{0.85\linewidth}
			\includegraphics[width=1\linewidth]{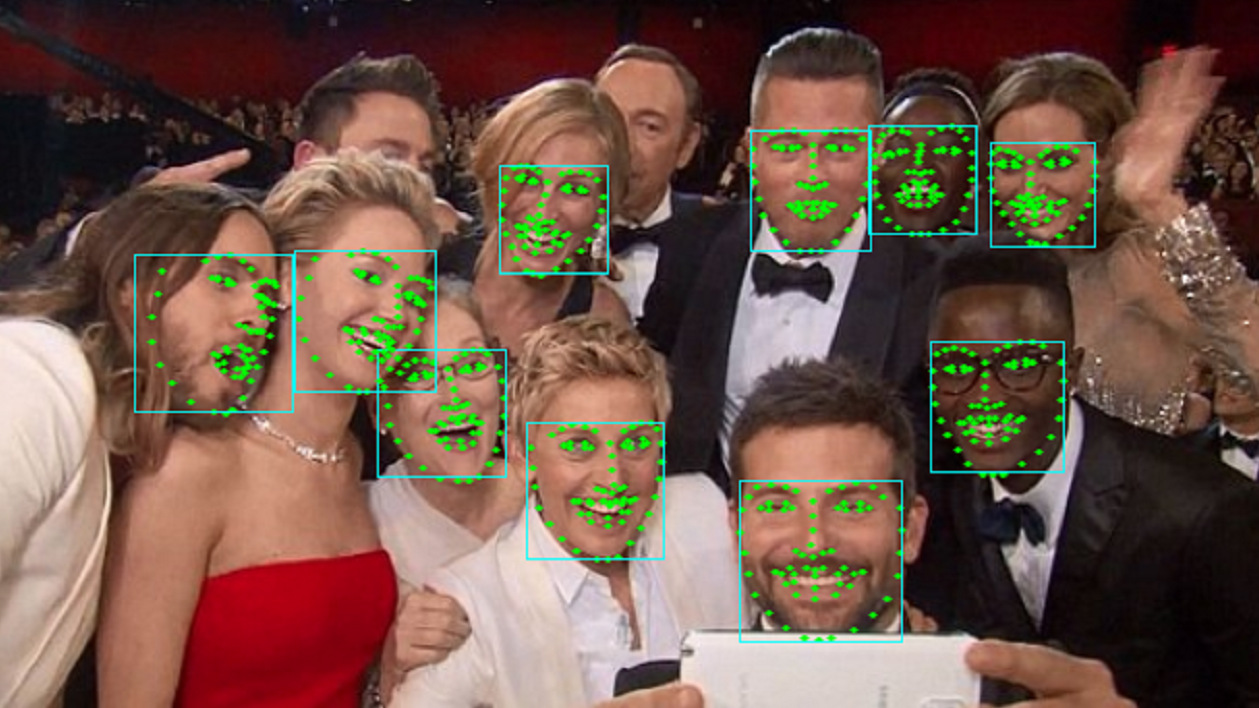}
			\setcounter{subfigure}{0}
		\end{subfigure}
	\end{center}
	\caption{Two examples with multiple faces. The faces are of different poses, expressions, and occlusions. Most of practical systems require to first detect faces, and then execute landmark localization on each detected face. Our practical system employs MTCNN to detect faces and PFLD 0.25X to localize landmarks, respectively.}
	\label{fig:res}
\end{figure*}

\textbf{Additional Results.} 
Figure \ref{fig:visual} displays a number of visual results of testing faces in 300W and AFLW. The faces are under different poses, lightings, expressions and occlusions, as well makeups and styles. Our PFLD 0.25X can obtain very pleasant landmark localization results. For the completeness of system, we simply employ MTCNN \cite{MTCNN} to detect faces in images/video frames, and then feed the detected faces into our PFLD to localize landmarks. In Fig. \ref{fig:res}, we give two example containing multiple faces. The results are obtained by our system. As can be seen, in the first case of Fig. \ref{fig:res}, all the faces are successfully detected, and the landmarks of each face are accurately localized. In the second picture, there are two faces in the back row missed, because they are severely occluded and hardly detected. {We emphasize that this omission comes from the face detector instead of the landmark detector.} The landmarks of all the detected faces are very well computed. \\ 

%

\begin{table}[t]
	\begin{center}
		\resizebox{0.48\textwidth}{!}{
			\begin{tabular}{ c|| c | c | c | c | c}
				\hline
				\textbf{Loss} & \textbf{$\ell_2$} & \textbf{$\ell_1$} &  \textbf{Ours w/o $\omega$} &  \textbf{Ours w/o $\theta$} & \textbf{Ours}\\
				\hline
				\hline
				\textbf{300W} (IPN) & 4.40 & 4.35 &  4.22 & 4.14 & \textbf{3.95}\\
				\hline
				\textbf{AFLW} (NME) & 2.33 & 2.31 &  2.23 & 2.16 & \textbf{1.88}\\
				\hline
			\end{tabular}
		}
	\end{center}
	
	\caption{Comparison of different loss functions.}
	\label{tab:abl}
\end{table}

%
%
%

\section{Concluding Remarks}

Three aspects of facial landmark detectors need to be concerned for being competent on large-scale and/or real-time tasks, which are accuracy, efficiency, and compactness. This paper proposed a practical facial landmark detector, termed as PFLD, which consists of two subnets, \textit{i.e.} the backbone network and the auxiliary network. The backbone network is built by the MobileNet blocks, which can largely release the computational pressure from convolutional layers, and make the model flexible in size according to a user's requirement by adjusting the width parameter. A multi-scale fully connected layer was introduced to enlarge the receptive field and thus enhance the ability of capturing face structures. To further regularize the landmark localization, we customized another branch, say the auxiliary network, by which the rotation information can be effectively estimated. Considering the geometric regularization and data imbalance issue, a novel loss was designed. The extensive experimental results demonstrate the superior performance of our design over the state-of-the-art methods in terms of accuracy, model size, and processing speed, therefore verifying that our PFLD 0.25X is a good trade-off for practical use.

In the current version, PFLD only adopts the rotation information  (yaw, roll and pitch angles) as the geometric constraint. It is expected to employ other geometric/structural information to help further improve the accuracy.  For instance, like LAB \cite{LAB}, we can regularize landmarks not to deviate far away from boundary lines. From another point of view, a possible attempt for boosting the performance is replacing the base loss, \textit{i.e.}, $\ell_2$ loss, by some task-specific ones. Designing more sophisticated weighting strategies in the loss would be also beneficial, especially when training data is imbalanced and limited. We leave the above mentioned thoughts as our future work.


{
\bibliographystyle{ieee}
\bibliography{aaai18ref}
}

\end{document}